\documentclass{article}

\usepackage[final,nonatbib]{nips_2017}

\usepackage[utf8]{inputenc} 
\usepackage[T1]{fontenc}    
\usepackage{hyperref}       
\usepackage{url}            
\usepackage{booktabs}       
\usepackage{amsfonts}       
\usepackage{nicefrac}       
\usepackage{microtype}      

\usepackage{todonotes}
\usepackage[inline]{enumitem}
\usepackage{subcaption}
\usepackage{graphicx}
\usepackage{booktabs}

\def\etal{\emph{et al}.\xspace}
\usepackage{xspace}
\newcommand{\Qbot}{\textsc{Q-bot}\xspace}
\newcommand{\Abot}{\textsc{A-bot}\xspace}

\title{Examining Cooperation in Visual Dialog Models}

\author{
Mircea~Mironenco  \thanks{Equal contribution by DK and MM. Order determined by coin toss.} \\
  University of Amsterdam\\
  Amsterdam, Netherlands \\
  \texttt{mircea.mironenco@student.uva.nl} \\ 
  \And
  Dana~Kianfar \footnotemark[1]\\
  University of Amsterdam\\
  Amsterdam, Netherlands \\
  \texttt{dana.kianfar@student.uva.nl} \\  
  \And
  Ke~Tran\\
  ILPS\\
  University of Amsterdam\\
  Amsterdam, Netherlands \\
  \texttt{m.k.tran@uva.nl} \\
   \And
   Evangelos~Kanoulas \\
  ILPS \\
  University of Amsterdam\\
  Amsterdam, Netherlands \\
  \texttt{e.kanoulas@uva.nl} \\
   \And
  Efstratios~Gavves \\
  QUVA Lab \\
  University of Amsterdam\\
  Amsterdam, Netherlands \\
  \texttt{e.gavves@uva.nl} \\
  }

\begin{document}
\maketitle

\begin{abstract} 

In this work we propose a blackbox intervention method for visual dialog models, with the aim of assessing the contribution of individual linguistic or visual components. Concretely, we conduct structured or randomized interventions that aim to impair an individual component of the model, and observe changes in task performance. We reproduce a state-of-the-art visual dialog model and demonstrate that our methodology yields surprising insights, namely that both dialog and image information have minimal contributions to task performance. The intervention method presented here can be applied as a sanity check for the strength and robustness of each component in visual dialog systems.

\end{abstract}

\section{Introduction} 
By combining vision and language, tasks that require high-level image understanding such as image captioning \cite{imgcap, imgcap2, imgcap3} and visual question answering \cite{vqa, vqa2, vqa3}, leverage the performance of deep neural networks in an attempt to simulate the way humans acquire and use information from different modalities in their environment. In more recent work, the availability of large-scale datasets has seen dialog models being proposed as a medium for communicating visual information \cite{datadialog, dialogcorp}. 

Two recent approaches have proposed models that aim to acquire natural language by multi-agent dialog on a downstream visual task \cite{visdialdataset, guesswhat}. Both models are asymmetrically \emph{primed} with initial textual and visual information, and leverage the information gap between two agents to simulate a human-like conversation. 

Our goal is to dissect the contributions of linguistic and visual components and their interplay. We believe that progressing visually-grounded conversational artificial intelligence requires the understanding of communicative protocols exchanged by the agents and how they utilize language and visual information cooperatively. In this work, we present a simple black-box \emph{intervention} method which aims to determine which source of linguistic and visual information agents exploit the most in order to complete the task. Our method is model-agnostic, and can be utilized as a sanity-check in the process of designing a new model. We demonstrate our method on the visual dialog model presented in \cite{visdial}. Our method empirically tests the robustness of the model with respect to variations in inputs, and performs unit testing on specific components.
\newpage

\section{Dialog Agents} 
\label{sec:models}
We replicate the supervised hierarchical recurrent encoder-decoder model proposed by Das~\etal~\cite{visdial} which engages in the cooperative image guessing game introduced in \cite{visdialdataset}. The game includes a question bot (\Qbot) and an answer bot (\Abot). Both bots are provided with a short description of an image and \Abot additionally accesses the image itself. Without seeing the image, \Qbot must communicate with \Abot by asking questions in order to guess the image. Since the outcome of the game is decided based on the accuracy of \Qbot's prediction of the ground truth image, \Abot must cooperate with \Qbot to win the game. Choosing this model was primarily motivated by its structure which incorporates multiple components working towards the same goal, as well as the VisDial dataset which is a sound testbed for this type of investigation due to its diversity and size.

To quantify the performance of the model, \Qbot is asked to rank the correct image among a set of candidates. Consistent with the original evaluation, we use mean-percentile rank (MPR) as our metric. A mean-percentile rank of 90\% means that the prediction of the \Qbot is closer to the ground truth image than 90\% of the images in the set. In the guessing game, forcing two agents to communicate through natural language enables humans to inspect the behaviours of the agents painlessly by creating meaningful and interpretable interventions.
\section{Methodology}
\label{sec:methodology}
We consider two types of interventions: (1) intervening on the initial condition (image and caption), and (2) intervening in the course of conversation by changing the responses generated by either agent. Understanding the initial condition is crucial in designing a conversational AI. For example, it sheds light on what data we should collect, or what is the natural interface between humans and AI. To understand the role of images and their captions, we perform interventions as follows.
\begin{itemize}
  \item \textbf{Image}: We replace image feature vector by random noise $z\sim \mathrm{Uniform}(0, 1)$. If the images are useful cues, then this intervention completely destroys a piece of essential information. We therefore expect a degradation in the evaluation performance.
  \item \textbf{Caption}: We replace a content word by a random word. Additionally we observe that many captions are poorly related to their corresponding images. 
\end{itemize}

Intervening during the dialogs on other hand, allows us to glimpse into the model's internal representations and its ability to exploit and exchange meaningful bits of information. We expect that intelligent systems should be sensitive to perturbations especially when they are not trained to cope against it. In this setting, we intervene on
\begin{itemize}
  \item \textbf{Question}: with probability $p$, each token in the question is replaced by a random token before giving it to  \Abot.
  \item \textbf{Answer}: with probability $p$, each token in the answer of \Abot  is replaced by a random token before giving it to  \Qbot.
\end{itemize}

In addition to random noise, we propose using \emph{negation} as a more principled approach to gaining insights into the cooperative behaviors of the two agents. Particularly we change the answer of \Abot from \emph{yes} to \emph{no} and vice versa. If the \Qbot behaves cooperatively its predictions should alternate dramatically. Here we choose to manipulate \Abot and observe the outcomes of the VisDial game because it is easier to negate the answer than the question. Moreover \emph{yes} and \emph{no} answers make up 37\% of the responses of \Abot in training and and 45\% in validation data, therefore it is reasonable to expect both bots to learn the negation concept.

\section{Experiments} 
\label{sec:experiments}

All experiments\footnote{Our code will be available at \url{https://github.com/danakianfar/Examining-Cooperation-in-VDM}} were performed on the model described in Das~\etal~\cite{visdial}. We perform the four interventions described in Section \ref{sec:methodology} during inference on the validation set. We study the performance of each modified dialog system, on the same task of ranking the image inside a collection, and compare it with the regular performance without interventions. Our aim is to understand whether the dialog or the image is being leveraged to provide further information, which the \Qbot can use to make better predictions. A priori, we expect to observe a large decline performance for all intervention experiments, as they essentially involve replacing an information source with random noise.

Specifically we intervene on the caption with different probabilities $p \in \{0.2, 0.4, 0.6, 0.8\}$ at the start of the dialog, and on the image, answers and questions at round 5 with $p=0.8$. In the case of the image intervention, we replace the entire image with random noise.

\section{Results} 
\label{sec:results}

\begin{figure}[ht!]
\centering
\begin{subfigure}[t]{0.48\textwidth}
    \includegraphics[width=1.01\textwidth]{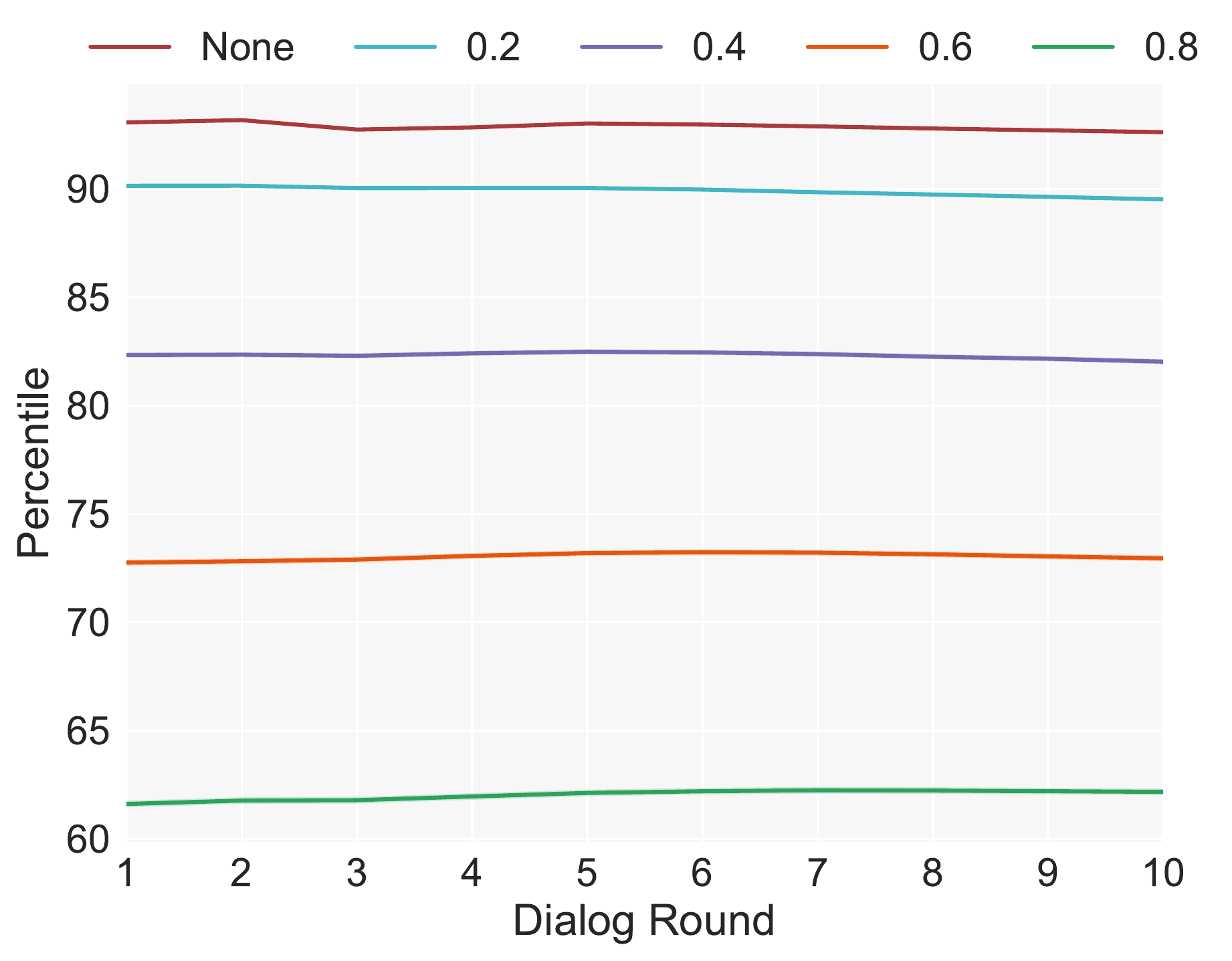}
    \caption{\textbf{Caption}: MPR when a token in the caption is replaced by another with $p \in \{0.2, 0.4, 0.6, 0.8\}$. "None" represents no interventions.}
    \label{fig:caption_intervention}
\end{subfigure} \hfill
\begin{subfigure}[t]{0.48\textwidth}
    \includegraphics[width=\textwidth]{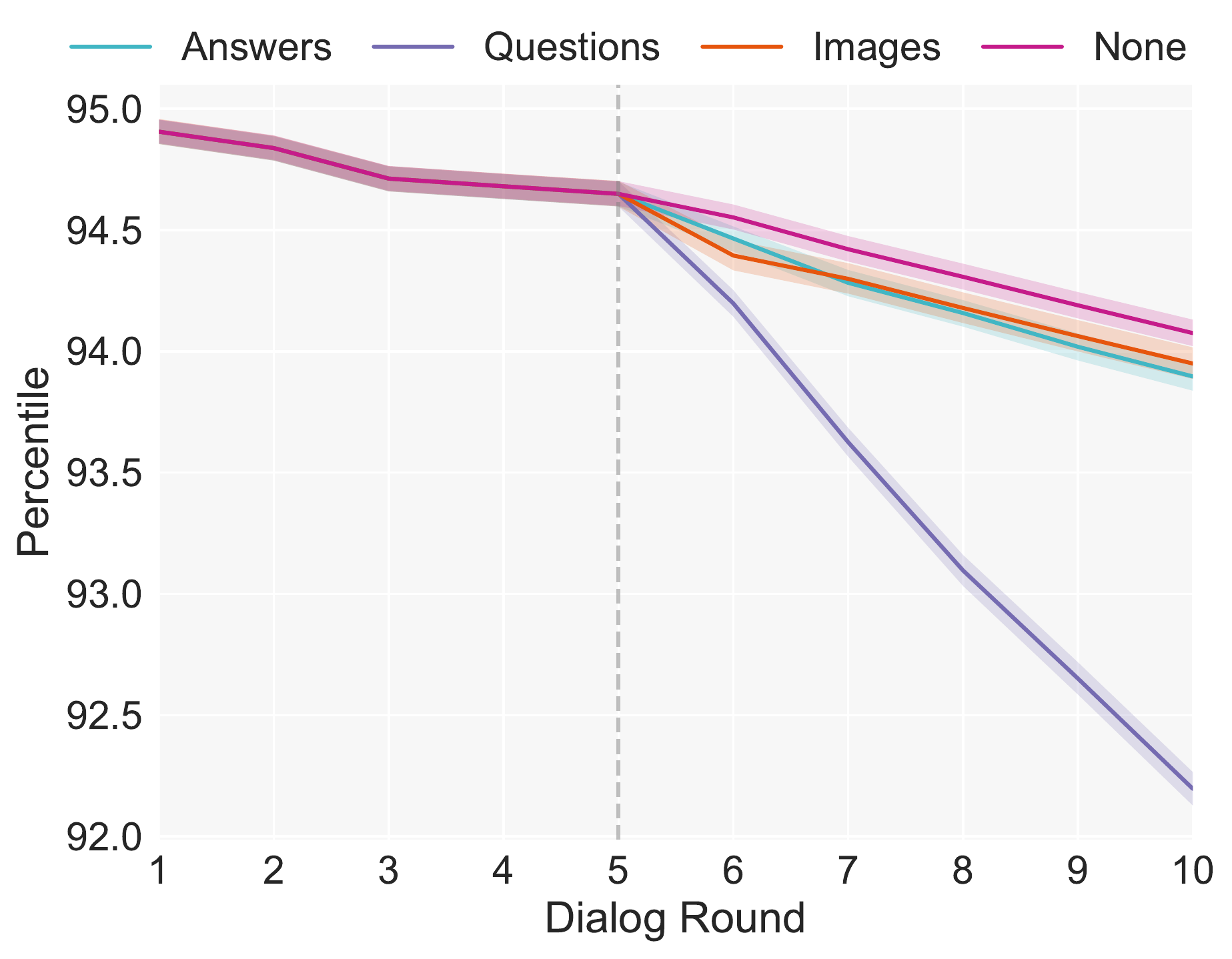}
    \caption{\textbf{Image, Question \& Answer}: MPR when we intervene starting at round 5 with $p=0.8$. "None" represents no interventions.}
    \label{fig:others_intervention}
\end{subfigure}
\caption{Comparison of rankings with and without interventions. Note that the y-axes in the two figures are not aligned.}
\label{fig:interventionresults}
\end{figure}

In this section we present the mean percentile rank (MPR) on the 40K validation set of the VisDial v0.9 dataset \cite{visdialdataset} for each intervention experiment (answers, captions, questions, and images) as well as the regular inference without interventions (called "None") as described in Section \ref{sec:methodology}. 
\paragraph{Caption Interventions} As displayed in Figure \ref{fig:caption_intervention} higher values of $p$ correspond to poorer performance. Despite the fact that the caption is only seen once at the start of the dialog, it nevertheless plays a very important role in the predictive performance of the network across all rounds. Figure \ref{fig:intervention_caption_appendix} in the appendix provides an example of a "positive" manual intervention where replacing the original image caption with a more informative one resulted in a much better ranking.
\paragraph{Image, Caption and Answer Interventions} As seen in Figure \ref{fig:others_intervention}, the interventions have a negative effect on the performance after the intervention on round 5. However, the decrease in performance by each individual experiment is much smaller than the decrease with interventions on captions. It suggests that \Qbot relies mainly on the caption  as it contains most of the information needed to make predictions. We also note that randomly intervening on the questions affects the performance the most.

We clearly observe a large discrepancy between rankings of caption interventions and the other experiments. Surprisingly, the decline in performance caused by interventions on answers and questions is less than expected, suggesting that the dialog itself is not used effectively for image identification. Interestingly replacing images with complete noise has minimal to no impact on the performance of the model. Q-BOT mainly relies on the caption provided at the beginning of the dialog to make predictions. This suggests that there is little cooperation between two bots. This effect is more clearly observed in the extreme case where we intervene with $p=1$ on all rounds, shown in Table \ref{tab:results}. We note that the ranking performance of the caption interventions is improved as the dialog continues, although it never recovers completely. 

\begin{table}[ht!]
\centering
\caption{MPR [\%] of each intervention experiment with $p=1.0$, where we intervened at each round during inference on the validation set. The difference between "None" and all other intervention rankings at the final round is shown in the final row.}
\begin{tabular}{@{} l c c c c c @{}}
\toprule
\textbf{Intervention} & None & Images & Captions & Answers & Questions  \\
\midrule
\textbf{Round 1} & 93.1 & 93.1 & 50.0 & 93.0 & 92.5\\
\textbf{Round 2} & 93.2 & 93.1 & 50.3 & 93.1 & 92.6\\
\textbf{Round 3} & 92.7 & 92.5 & 50.7 & 92.5 & 91.8\\
\textbf{Round 4} & 92.8 & 92.5 & 51.3 & 92.5 & 91.4\\
\textbf{Round 5} & 93.0 & 92.7 & 51.6 & 92.5 & 91.3\\
\textbf{Round 6} & 93.0 & 92.6 & 51.9 & 92.4 & 90.9\\
\textbf{Round 7} & 92.9 & 92.5 & 52.2 & 92.2 & 90.6\\
\textbf{Round 8} & 92.8 & 92.4 & 52.3 & 92.0 & 90.2\\
\textbf{Round 9} & 92.7 & 92.3 & 52.4 & 91.9 & 89.9\\
\textbf{Round 10} & 92.6 & 92.2 & 52.5 & 91.7 & 89.6\\
\midrule
\textbf{Gap @10} & \textbf{0.0} & \textbf{0.4} & \textbf{40.1} & \textbf{0.8} & \textbf{3.0}\\
\bottomrule
\end{tabular}
\label{tab:results}
\end{table}

\begin{table}[ht!]
\centering
\caption{MPR [\%] of Negation intervention results.}
\begin{tabular}{@{} c c c c c c c c c c c @{}}
\toprule
\textbf{Round} & 1 & 2 & 3 & 4 & 5 & 6 & 7 & 8 & 9 & 10  \\
\midrule
\textbf{8} &   &   &   &   &   &   &   & 94.3 & 94.1 & 93.9 \\
\textbf{6} &  &   &   &   &   & 94.6 & 94.4 & 94.2 & 94.1 & 93.9 \\
\textbf{4} &   &   &   & 94.7 & 94.6 & 94.5 & 94.3 & 94.2 & 94.0 & 93.8 \\
\textbf{2} &  & 94.8 & 94.7 & 94.6 & 94.6 & 94.4 & 94.3 & 94.1 & 94.0 & 93.8 \\
\textbf{0} & 94.8 & 94.8 & 94.7 & 94.6 & 94.5 & 94.4 & 94.2 & 94.1 & 93.9 & 93.7 \\
\bottomrule
\end{tabular}
\label{tab:neg_results}
\end{table}

Table~\ref{tab:neg_results} shows the results of negation intervention experiments. We start intervening from round 0, 2, 4, 6, and 8. Comparing the MPR across all rounds (columns in the table), we see that \Qbot is sensitive to the non-cooperative behaviors of \Abot, however only to a small degree ($<$ 0.3\%). Our results suggest that in addition to downstream evaluation in a cooperative game setting, forcing one agent to play non-cooperatively could help researchers design a better experimental setup to understand the cooperative behaviours amongst agents in the system. 

\section{Discussion} 
\label{sec:discussion}

We have presented a simple yet effective method for assessing the interaction of linguistic and visual components in visual dialog models. Using an example which combines multiple sources of information in a cooperative multi-agent setup, we have demonstrated that impairing individual components can reveal the extent to which each information source is exploited by the agents to accomplish their goals. A pitfall with designing multi-modal and blackbox systems is that the role of individual components can not be deduced from the overall performance of the model.

In a series of surprising results, we discovered that the role of images is minimal in the image retrieval performance. Furthermore, our evaluation suggests that the dialog itself was not exploited significantly by the bots in the cooperative setting. We argue that designing multi-modal systems requires careful evaluation, or \textit{unit testing}, of each component. Grounding natural language is a difficult problem as models which combine modalities must account for the individual impact of each information source.

We encourage future researchers to account for the effect of each modality on the total performance of the model. An interesting research direction is how to learn interaction models of independent modalities which can be shown to generalize well in conjunction, and avoid the pitfall of overfitting to spurious correlations that optimize the surrogate learning objectives of each independent modality. Reinforcement learning (RL) has traditionally been used to train dialog models and has been motivated as a natural training paradigm for dialogue models, and has been applied to visual dialog models in \cite{visdial,guesswhat}. We did not include RL methods in our analysis and leave such assessment to future work.

\section{Acknowledgements}

This research was supported by the Google Faculty Research Award program and the Dutch national program COMMIT.

\bibliography{ref}

\newpage 
\section{Appendix}

\subsection{Manual interventions}
We present examples of manual interventions below. All intervened values are displayed in blue. The original inference is displayed on the left column of each example, and the intervened dialog is presented on the right column. The normal inference rankings are displayed in dark blue and the intervened ranking in light blue.

\begin{figure}[ht!]
    \centering
    \includegraphics[width=\textwidth]{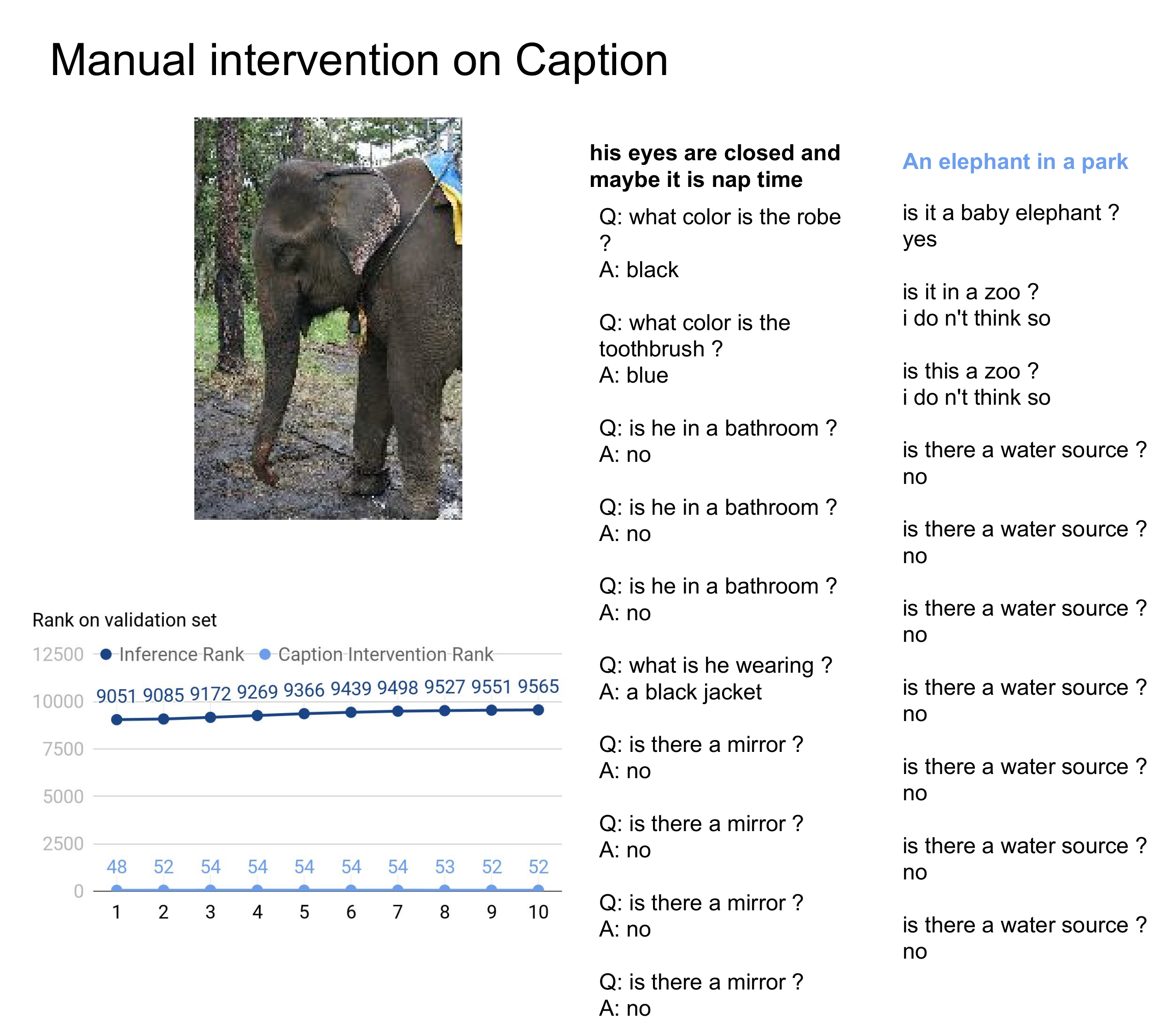}
    \caption{Manual 'positive' intervention on the caption of the image. The original caption (top of left column in bold) is uninformative and results in a poor rankings of approximately 9000 out of approximately 40000. Changing the caption to make it more descriptive (right column in blue) improves the rankings dramatically to around 50.}
    \label{fig:intervention_caption_appendix}
\end{figure}

\begin{figure}[ht!]
    \centering
    \includegraphics[width=\textwidth]{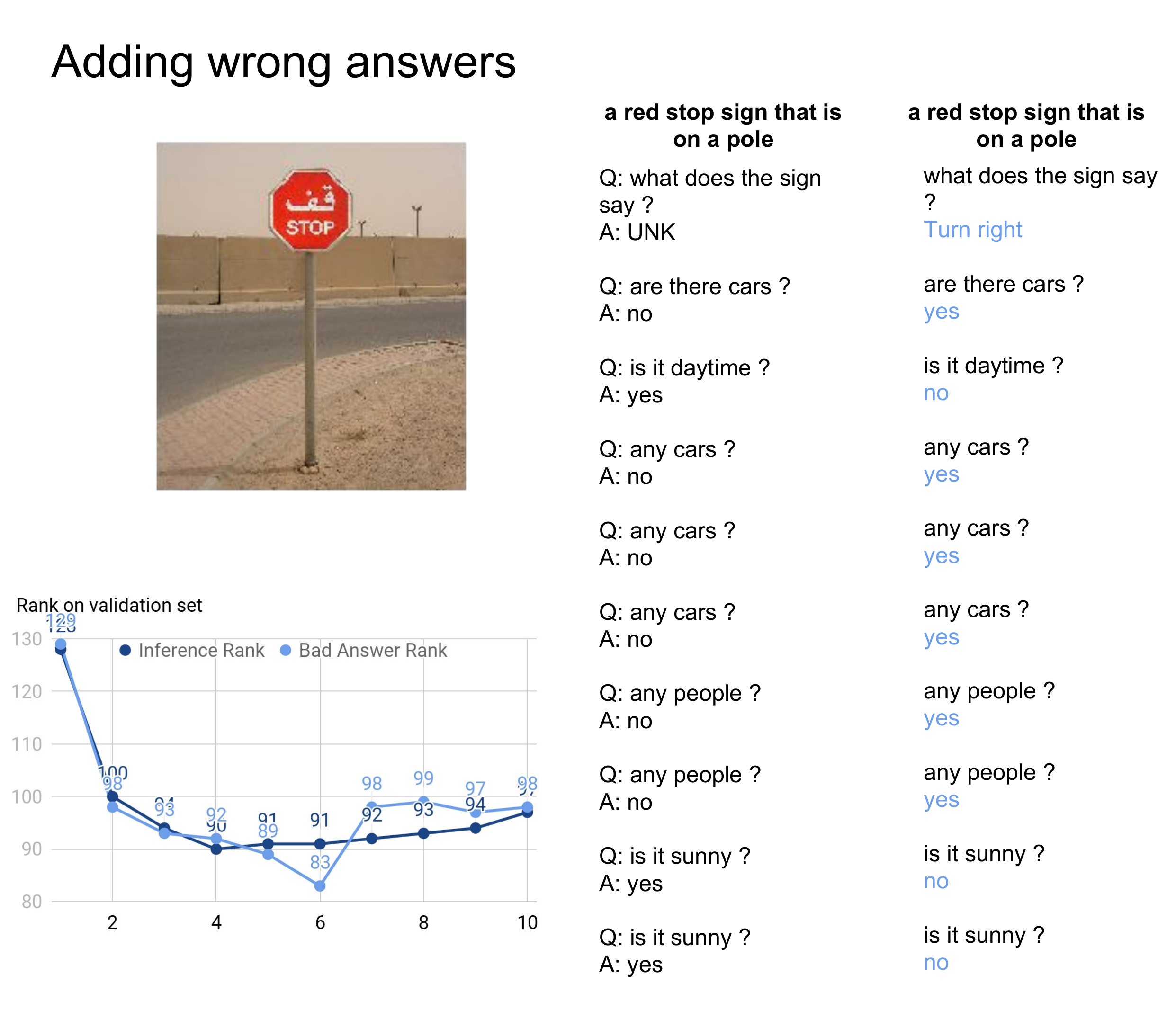}
    \caption{Manual intervention on the answers. We consistently provide answers that are either false or which negate the inferred dialog (displayed in blue). The original dialog achieves a final ranking of 97 out of approximately 40000. Surprisingly, the interventions do not cause a very large perturbation despite being misleading and uninformative.}
    \label{fig:intervention_answers_appendix}
\end{figure}

\begin{figure}[ht!]
    \centering
    \includegraphics[width=\textwidth]{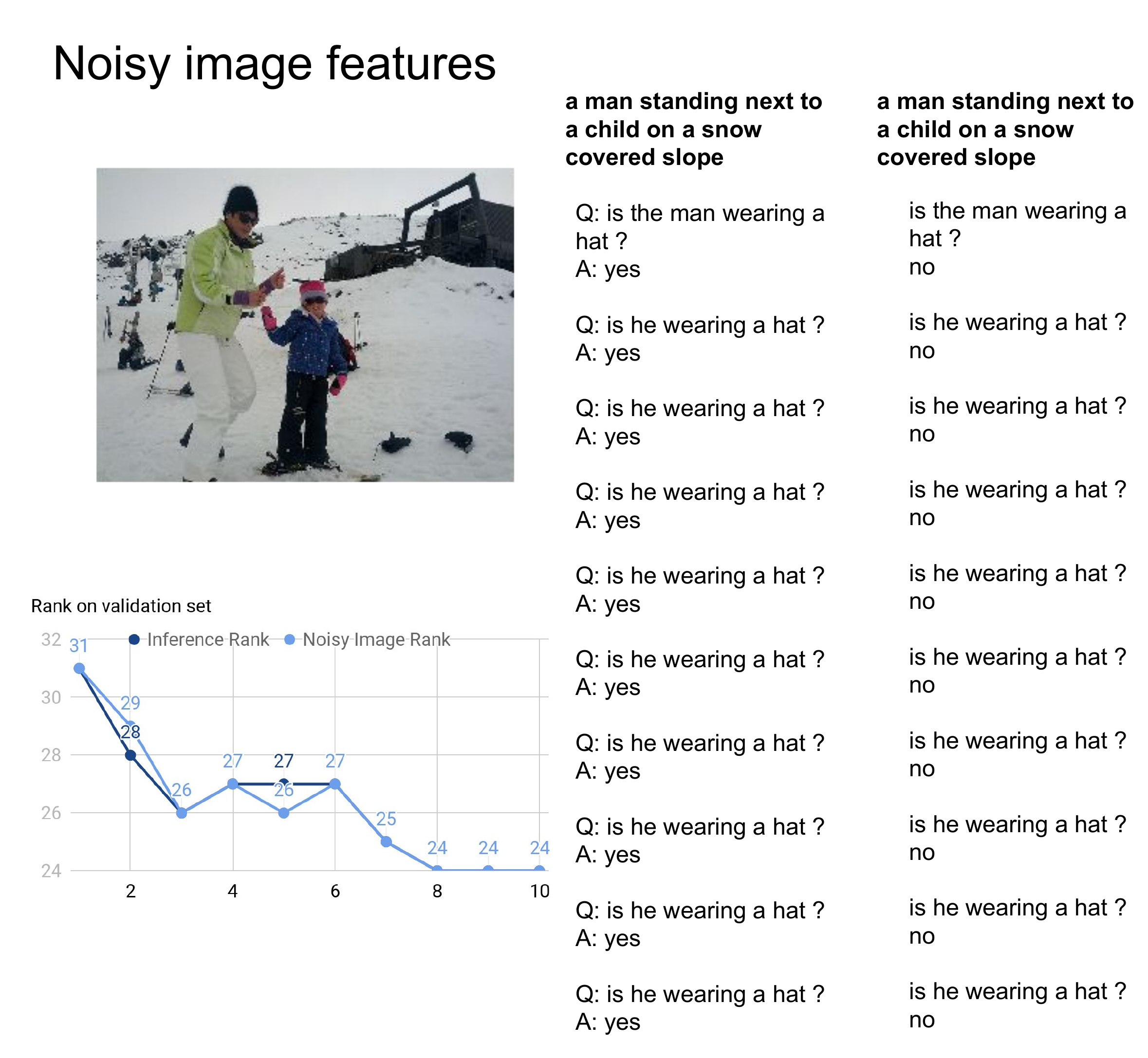}
    \caption{Manual intervention where we replaced the image feature vector with random noise. Surprisingly, there is no noticeable change in image rankings. The decoded sequences by the answer bot are however different. }
    \label{fig:intervention_image_appendix}
\end{figure}

\begin{figure}[ht!]
    \centering
    \includegraphics[width=\textwidth]{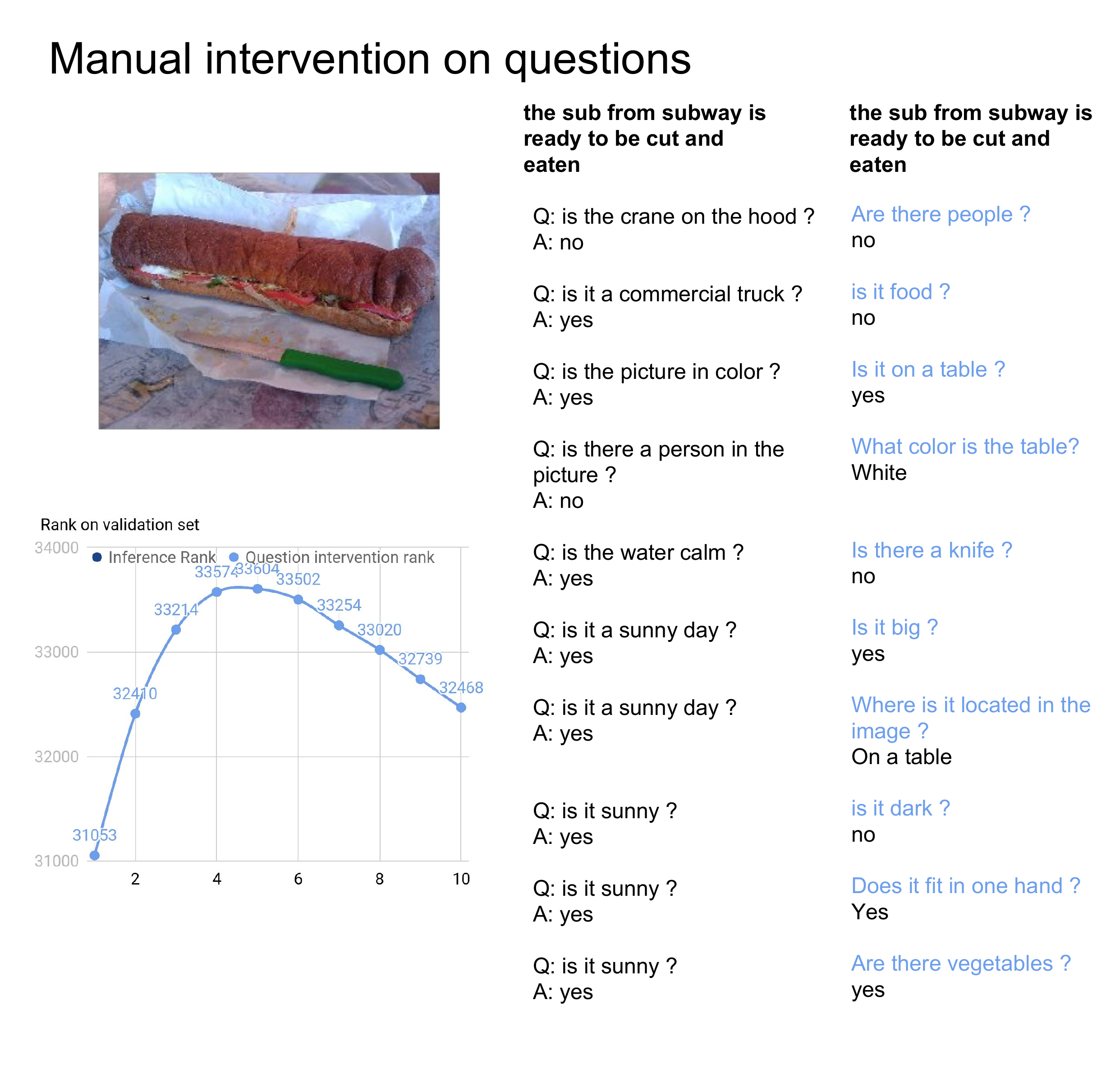}
    \caption{Manual intervention where we provided more meaningful questions (displayed in blue). Surprisingly, there is no change in image rankings. The two image rankings perfectly overlap.}
    \label{fig:intervention_question_appendix}
\end{figure}

\end{document}